\documentclass[10pt,twocolumn,letterpaper]{article}

\usepackage{cvpr}
\usepackage{times}
\usepackage{epsfig}
\usepackage{graphicx}
\usepackage{amsmath}
\usepackage{amssymb}
\usepackage{authblk}
\usepackage{booktabs}

\usepackage{multirow}
\usepackage{caption}
\usepackage{array}
\usepackage{subfigure}

\usepackage{enumitem}
\usepackage{color}

\usepackage[pagebackref=true,breaklinks=true,letterpaper=true,colorlinks,bookmarks=false]{hyperref}

\cvprfinalcopy 

\def\cvprPaperID{5306} 

\begin{document}

\title{MemNAS: Memory-Efficient Neural Architecture Search with Grow-Trim Learning}

\author{
Peiye Liu$^{\dag\S}$, Bo Wu$^{\S}$, Huadong Ma$^{\dag}$, and Mingoo Seok$^{\S}$\\
$^\dag$Beijing Key Lab of Intelligent Telecommunication Software and Multimedia,\\
Beijing University of Posts and Telecommunications, Beijing, China\\
$^\S$Department of Electrical Engineering, Columbia University, NY, USA\\
{\tt\small \{liupeiye, mhd\}@bupt.edu.cn}
{\tt\small \{bo.wu, ms4415\}@columbia.edu}
}
\maketitle
\thispagestyle{empty}

\begin{abstract}
Recent studies on automatic neural architecture search techniques have demonstrated significant performance, competitive to or even better than hand-crafted neural architectures. However, most of the existing search approaches tend to use residual structures and a concatenation connection between shallow and deep features. A resulted neural network model, therefore, is non-trivial for resource-constraint devices to execute since such a model requires large memory to store network parameters and intermediate feature maps along with excessive computing complexity.
To address this challenge, we propose \textbf{MemNAS}, a novel growing and trimming based neural architecture search framework that optimizes not only performance but also memory requirement of an inference network.
Specifically, in the search process, we consider running memory use, including network parameters and the essential intermediate feature maps memory requirement, as an optimization objective along with performance.
Besides, to improve the accuracy of the search, we extract the correlation information among multiple candidate architectures to \textit{rank} them and then choose the candidates with desired performance and memory efficiency.
On the ImageNet classification task, our MemNAS achieves 75.4\% accuracy, 0.7\% higher than MobileNetV2 with 42.1\% less memory requirement. Additional experiments confirm that the proposed MemNAS can perform well across the different targets of the trade-off between accuracy and memory consumption.

\end{abstract}
\begin{figure*} 
   \centering
   \includegraphics[width=1\linewidth]{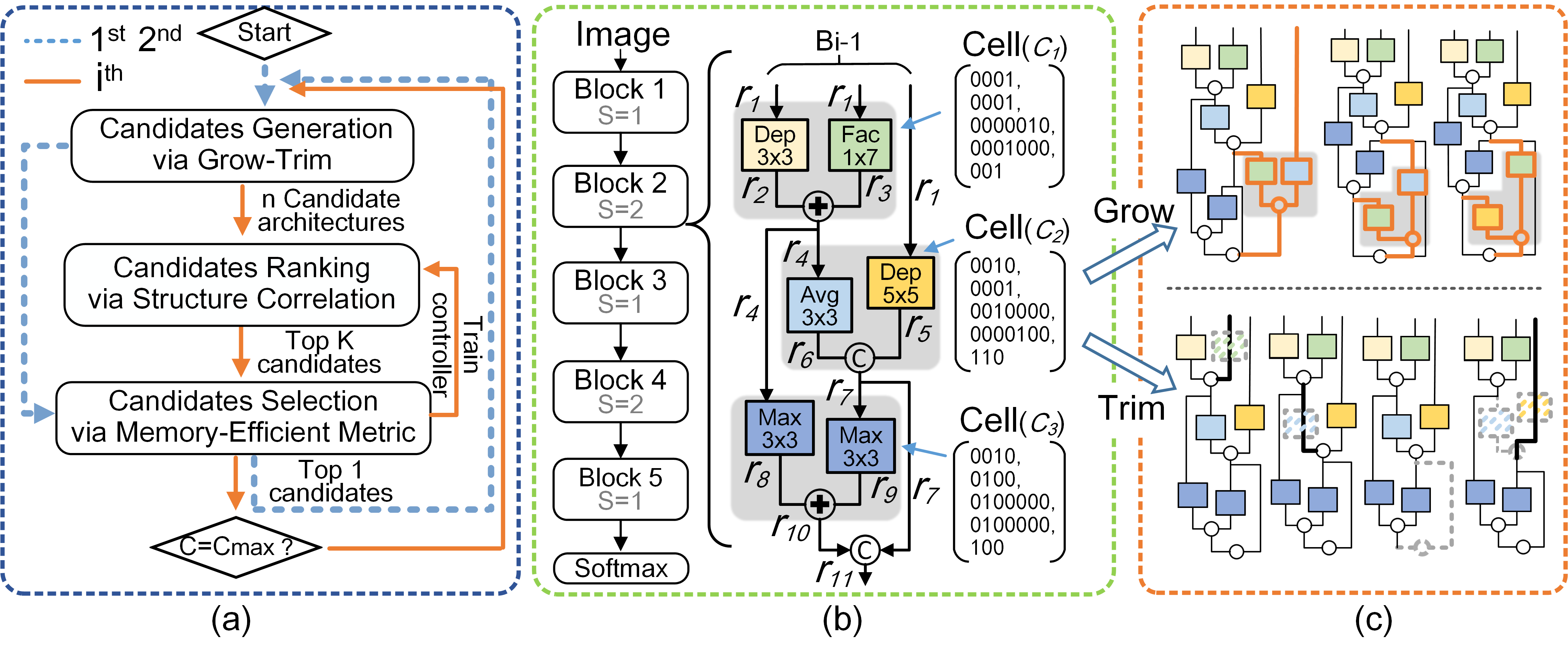}
   \vspace{-0.8cm}
	\caption{
	\textbf{ (a) Flow Chart of the Proposed MemNAS}. It has mainly three steps: i) candidate neural network generation, ii) the top-$k$ candidate generation using the proposed structure correlation controller, and iii) candidate training and selection.
	\textbf{(b) The Network Structure for CIFAR-10.} The neural network architecture has five blocks. Each block contains several cells with stride (S) 1 and 2. Each cell ($C_{i}$), shown in the gray background, is represented by a tuple of five binary vectors. $r_{i}$ represents the intermediate representations in one block.
	\textbf{(c) Examples of Candidates by Growing and Trimming a base Network.} The cells in the gray background are newly added. The layers with the dashed outlines are removed. We remove only one layer or one edge only in a block when we trim a neural network. But we add the same cell to all five blocks when we grow in CIFAR-10.
	}
    \vspace{-0.4cm}
	\label{fig:memnet}
\end{figure*}
\section{Introduction}

Deep Neural Networks (DNNs) have demonstrated the state-of-the-art results in multiple applications including classification, search, and detection \cite{he2016deep,real2017large, liu2016ssd, liu2017beyond,liu2018third,redmon2016you,liu2016deep}.
However, those state-of-the-art neural networks are extremely deep and also highly complicated, making it a non-trivial task to hand-craft one. This has drawn researchers' attention to the neural architecture search (NAS), which involves the techniques to construct neural networks without profound domain knowledge and hand-crafting \cite{tan2019mnasnet,cai2018efficient,zoph2016neural,zoph2018learning}.

On the other hand, whether to design a neural architecture manually or automatically, it becomes increasingly important to consider the target platform that performs inference.
Today, we consider mainly two platforms, a data center, and a mobile device.
A neural network running on a data center can leverage massive computing resources. Therefore, the NAS works for a data center platform focus on optimizing the speed of the search process and the performance (accuracy) of an inference neural network \cite{cai2018efficient,zoph2016neural,zoph2018learning,hammerla2016deep,cai2019once,yu2019autoslim}.
A mobile computing platform, however, has much less memory and energy resources.
Hence, NAS works for a mobile platform have attempted to use lightweight network layers for reducing memory requirement \cite{hsu2018monas,dong2018ppp-net:,michel2019dvolver}. Besides, off-chip memory access such as FLASH and DRAM is 3 or 4 orders of magnitudes power-hungrier and slower than on-chip memory \cite{wulf1995hitting}.
Therefore, it is highly preferred to minimize network size to the level that the network can fit entirely in the on-chip memory of mobile hardware which is a few MB.

Unfortunately, most of the existing NAS approach, whether based on reinforcement learning (RL) or evolutionary algorithm (EA), adopts a \textit{grow}-only strategy for generating new network candidates.  
Specifically, in each search round, they add more layers and edges to a base architecture, resulting in a network that uses increasingly more memory and computational resources. 

Instead, we first propose a \textit{grow-and-trim} strategy in generating candidates in NAS, where we can remove layers and edges during the search process from the base architecture without significantly affecting performance.
As compared to the grow-only approaches, the proposed grow-and-trim approach can generate a large number of candidate architectures of diverse characteristics, increasing the chance to find a network that is high-performance and memory-efficient.

Such a large number of candidate architectures, however, can be potentially problematic if we do not have an accurate method to steer the search and thus choose the desired architecture.
To address this challenge, we propose a \textit{structure correlation controller} and a \textit{memory-efficiency metric}, with which we can accurately choose the best architecture in each search round.
Specifically, the structure correlation controller extracts the \textit{relative} information of multiple candidate network architectures, and by using that information it can estimate the \textit{ranking} of candidate architectures.
Besides, the memory-efficiency metric is the weighted sum of the accuracy performance of a network and the memory requirement to perform inference with that network.

We perform a series of experiments and demonstrate that MemNAS can construct a neural network with competitive performance yet less memory requirement than the state of the arts. The contributions of this work are as follows:
\vspace{-0.2cm}
\begin{itemize}
\setlength{\itemsep}{-0.5ex}
\item We propose a neural architecture search framework (MemNAS) that grows and trims networks for automatically constructing a memory-efficient and high-performance architecture.
\item We design a structure correlation controller to predict the ranking of candidate networks, which enables MemNAS effectively to search the best network in a larger and more diverse search space. 
\item We propose a memory-efficiency metric that defines the balance of accuracy and memory requirement, with which we can train the controller and evaluate the neural networks in the search process. The metric considers the memory requirement of both parameters and essential intermediate representations.
To estimate the memory requirement without the details of a target hardware platform, we also develop a lifetime-based technique which can calculate the upper bound of memory consumption of an inference operation.
\end{itemize}

\section{Related Work}
\subsection{Hand-Crafted Neural Architecture Design}
It has gained a significant amount of attention to perform inference with a high-quality DNN model on a resource-constrained mobile device\cite{liu2018ktgan,szegedy2015going,iandola2016squeezenet, howard2017mobilenets,sandler2018mobilenetv2, li2017pruning}. This has motivated a number of studies to attempt to scale the size and computational complexity of a DNN without compromising accuracy performance. In this thread of works, multiple groups have explored the use of filters with small kernel size and concatenated several of them to emulate a large filter. For example, GoogLeNet adopts one $1 \times N$ and one $N \times 1$ convolutions to replace $N \times N$ convolution, where $N$ is the kernel size \cite{szegedy2015going}. Similarly, it is also proposed to decompose a 3-D convolution to a set of 2-D convolutions. For example, MobileNet decomposes the original $N \times N \times M$ convolution ($M$ is the filter number) to one $N \times N \times 1$ convolution and one $1 \times 1 \times M$ convolution  \cite{howard2017mobilenets}. This can reduce the filter-related computation complexity from $N\times N\times M\times I\times O$ ($I$ is the number of input channels and $O$ is the number of output channels) to $N \times N \times M \times O+M \times I\times O$. In addition, SqueezeNet adopts a fire module that squeezes the network with $1 \times 1$ convolution filters and then expands it with multiple $1 \times 1$ and $3 \times 3$ convolution filters \cite{iandola2016squeezenet}. ShuffleNet utilizes the point-wise group convolution to replace the $1 \times 1$ filter for further reducing computation complexity \cite{zhang2018shufflenet}.

\subsection{Neural Architecture Search}
Recently, multiple groups have proposed neural architecture search (NAS) techniques which can automatically create a high-performance neural network. Zoph \textit{et} al. presented a seminal work in this area, where they introduced the reinforcement learning (RL) for NAS \cite{zoph2016neural}.
Since then, several works have proposed different NAS techniques.
Dong \textit{et} al. proposed the DPP-Net framework \cite{dong2018ppp-net:}.
The framework considers both the time cost and accuracy of an inference network.
It formulates the down-selection of neural network candidates into a multi-objective optimization problem \cite{hochman1969pareto} and chooses the top-\textit{k} neural architectures in the Pareto front area.
However, the framework adopts CondenseNet \cite{huang2018condensenet} which tends to produce a large amount of intermediate data.
It also requires the human intervention of picking the top networks from the selected Pareto front area in each search round.
Hsu \textit{et} al. \cite{hsu2018monas} proposed MONAS framework, which employs the reward function of prediction accuracy and power consumption. While it successfully constructs a low-power neural architecture, it considers only a small set of existing neural networks in its search, namely AlexNet \cite{krizhevsky2012imagenet}, CondenseNet \cite{huang2018condensenet}, and their variants. Michel \textit{et} al. proposed the DVOLVER framework \cite{michel2019dvolver}. However, it only focuses on the minimization of network parameters along with the performance. Without considering intermediate representation, DVOLVER may produce an inference network still requiring a large memory resource.

\begin{figure}[t]
\centering
\includegraphics[width=1\linewidth]{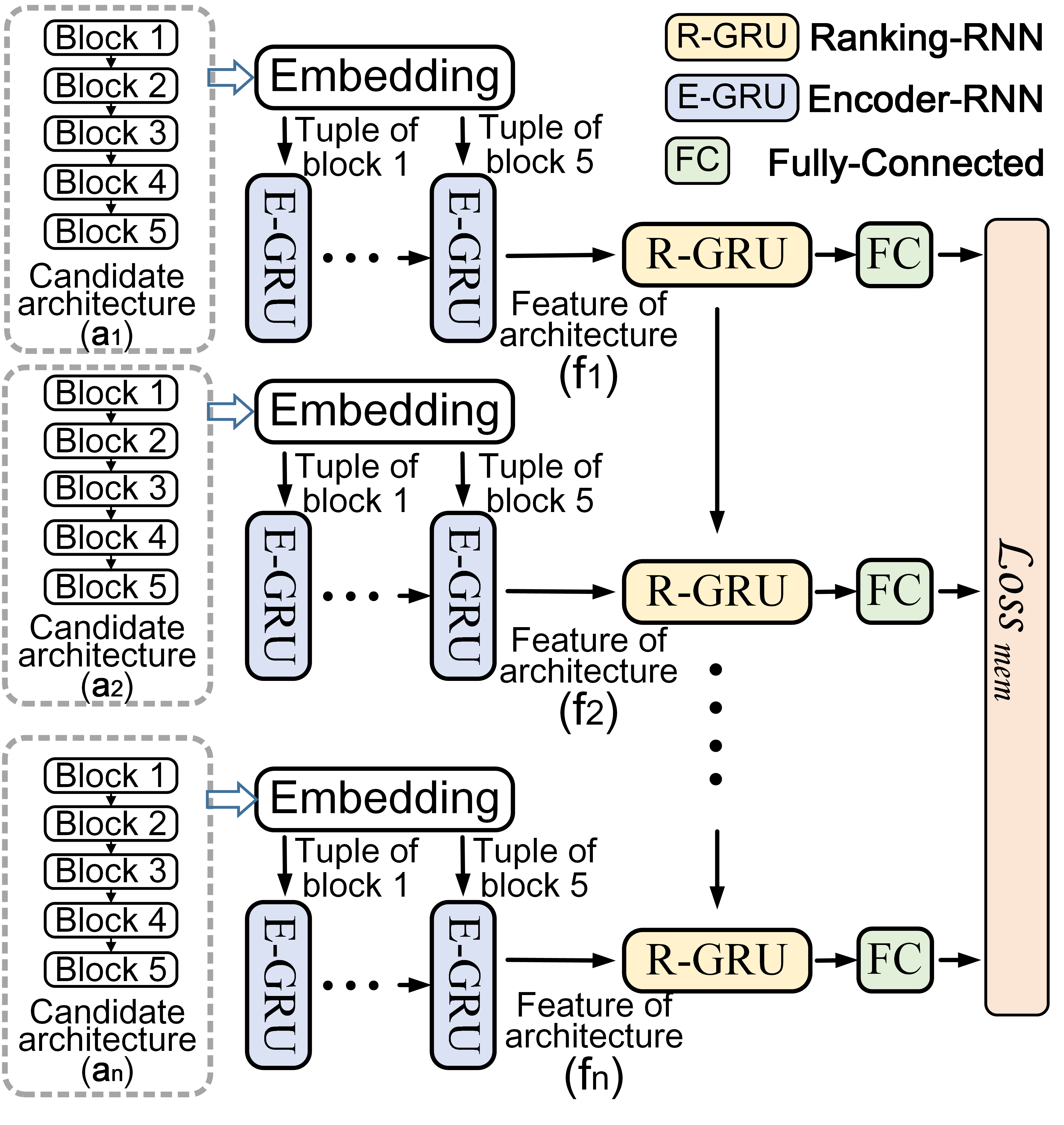}
\vspace{-0.8cm}
	\caption{
	\textbf{The Proposed Structure Correlation Controller (SCC).} All the candidate architectures ($a_1$, $a_2$, .. $a_n$) are mapped to the features ($f_1$, $f_2$, ... $f_n$).}
\vspace{-0.4cm}
\label{fig:rr}
\end{figure}
\section{Proposed Method}
\subsection{Overview}

The goal of MemNAS is to construct a neural architecture that achieves the target trade-off between inference accuracy and memory requirement. Figure~\ref{fig:memnet} (a) depicts the typical search process consisting of multiple rounds. In each round, first, it generates several candidate architectures via the grow and trim technique. Second, it ranks the candidate architectures, using the structure correlation controller, in terms of the memory-efficiency metric, resulting in top-$k$ candidates. Third, we train the top-$k$ candidates and evaluate them in terms of the memory-efficiency metric. The best architecture is chosen for the next search round. Finally, we train the controller using the data we collected during the training of the top-$k$ candidates.

Figure~\ref{fig:memnet} (b) shows the template architecture of the neural networks used in the MemNAS. It has five series-connected \textit{blocks}. Each block consists of multiple \textit{cells}. Each cell has two operation \textit{layers} in parallel and one layer that sums or concatenates the outputs of the operation layers.

The location, connections, and layer types (contents) of a cell are identified by a tuple of five vectors, $(I_{1},I_{2},L_{1},L_{2},O)$. In a tuple, $I_{1}$ and $I_{2}$ are one hot encoded binary vector that represents the two inputs of a cell. For example, as shown in Figure~\ref{fig:memnet}(b) top right, the two inputs of the $C_{1}$ are both $r_1$ (=0001). Thus, the tuple's first two vectors are both $r_1$. Similarly, the second cell $C_{2}$ in Figure~\ref{fig:memnet}(b) mid-right has two inputs, $r_4$ (0010) and $r_1$ (0001). On the other hand, $O$ represents the type of the combining layer, namely 001: summing two operation layers but the output not included in the final output of the block; 110: concatenating two operation layers and the output included in the final output of the cell. $L_1$ and $L_2$ represent the types of two operation layers in a cell. They are also one-hot encoded.

A cell employs two operation layers from a total of seven operation layers. The two layers can perform the same operation. The seven operation layers and their binary vectors identifier are:
\vspace{-0.2cm}
\begin{itemize}[leftmargin=20pt]
\setlength{\itemsep}{-0.5ex}
    \item 3 x 3 convolution (0000001)
    \item 3 x 3 depth-wise convolution (0000010)
    \item 5 x 5 depth-wise convolution (0000100)
    \item 1 x 7 followed by 7 x 1 convolution (0001000)
    \item 3 x 3 average pooling (0010000)
    \item 3 x 3 max pooling (0100000)
    \item 3 x 3 dilated convolution (1000000)
\end{itemize}
 \vspace{-0.1cm}
\noindent These layers are designed for replacing conventional convolution layers that require large memory for buffering intermediate representation \cite{liu2018progressive}. The stride of layers is defined on a block-by-block basis. If a block needs to maintain the size of feature maps, it uses the stride of 1 (see the first block in Figure~\ref{fig:memnet}(b)). To reduce the feature map size by half, a block can use the stride of 2.

Inspired by the evolutionary algorithm \cite{real2017large}, MemNAS adds a new cell to each of the blocks in the same location. Besides, MemNAS removes layers differently in each block.

\subsection{Grow-and-Trim Candidate Generation}
\label{sec:at}
In MemNAS, each round begins with generating a large number of neural network candidates based on the network chosen in the previous round (called a base network). The collection of these generated candidate architectures constructs the search space of the round. It is important to make the search space to contain diverse candidate architectures. This is because a large search space can potentially increase the chance of finding the optimal network architecture that meets the target.

We first generate new candidates by \textit{growing} a base network. Specifically, we add a new cell to all of the five blocks in the same way.
We also generate more candidates by \textit{trimming} a base network. We consider two types of trimming. First, we can replace one of the existing operation layers with an identity operation layer. Second, we can remove an edge. If the removal of an edge makes a layer to lose its input edge or a cell's output to feed no other cells, we remove the layer and the cell (see Figure~\ref{fig:memnet}(c) bottom, the second last Trim Generation example). Note that we perform trimming in \textit{only one} of the five blocks once.

The size of the search space of all possible candidates via growing can be formulated to:
\begin{equation}
|S_{g}|=|I|^{2}*|L|^{2}*|C|^{2},
\end{equation} 
where $I$ denotes the number of available input locations in a cell, $L$ represents the number of available operation layer types, and $C$ denotes the number of connection methods. 
On the other hand, the size of the search space of all possible candidates via trimming can be formulated to:
\begin{equation}
|S_{t}|=\sum_{i=1}^{B}(l_{i}+c_{i}+e_{i}),
\end{equation}
where $B$ is the number of blocks, $l_{i}$ is the number of the layers in block $i$, $c_{i}$ is the number of the cells in block $i$, and $e_{i}$ is the number of the existing outputs in the final concatenation of block $i$. 

\begin{figure}[t]
\centering
\includegraphics[width=0.9\linewidth]{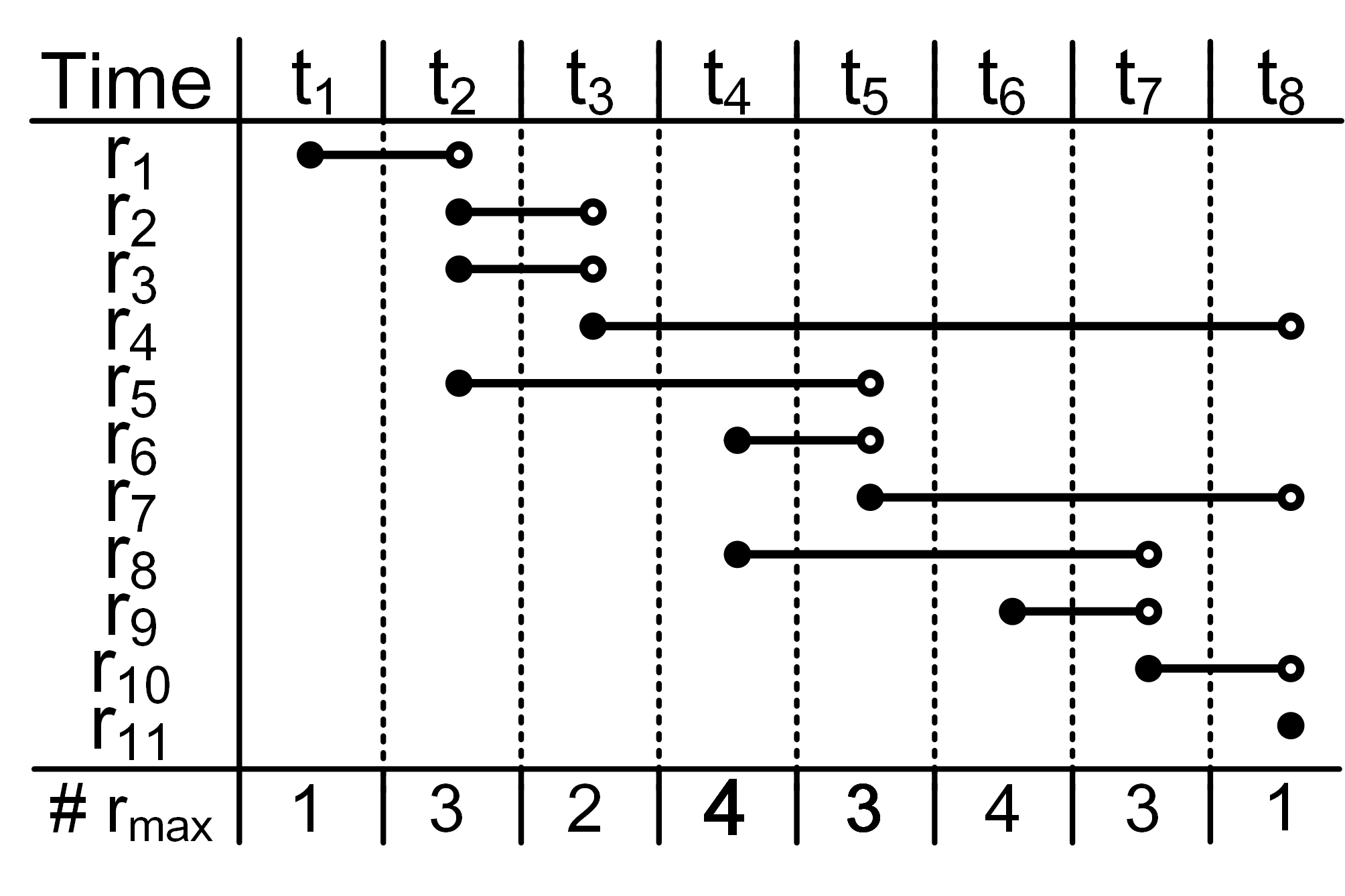}
 \vspace{-0.2cm}
\caption{\textbf{An Example Lifetime Plot.} We draw the lifetime plot for the neural network block architecture in Figure~\ref{fig:memnet}(b). The solid circle denotes the generation of intermediate representations and the hallow circuits denote the deletion of intermediate data. For simplicity, we assume the data size of each intermediate representation ($r_i\in r_1, r_2, ... r_{10}$) is 1. The last row represents the memory requirement of each time; the largest among them determines the memory requirement of hardware for intermediate data representation. }
\label{fig:lt}
\vspace{-0.2cm}
\end{figure}
\subsection{Structure Correlation Controller}
\label{sec:rr}
Our grow-and-trim technique enables MemNAS to explore a large search space containing a diverse set of neural network architectures. Particularly, we trim an individual layer or edge on a block-by-block basis, largely increasing the diversity and size of the search space.
To find the optimal neural network architecture without training all the candidates, therefore, it is critical to build a controller (or a predictor) for accurately finding the top-$k$ candidates.

To this goal, we propose a structure correlation controller (SCC).
This controller can map the stacked blocks of each candidate to a feature and then estimate the \textit{ranking} of the candidate networks in terms of the user-specified target of accuracy and memory requirement.
The SCC aims to extract the relative information among the candidate networks to evaluate a \textit{relative} performance, which is more accurate than the existing controllers \cite{michel2019dvolver} predicting the \textit{absolute} score of each candidate network individually and then rank the top-$k$ based on the absolute scores.

The SCC consists of mainly two recursive neural network (RNN) layers: i) the \textit{encoding layer} to map the blocks of each candidate to a feature and ii) the \textit{ranking layer} to map the features of all the candidates to the ranking score (Figure~\ref{fig:rr}).
In the encoding layer, we first feed candidate networks to the embedding layer, obtaining a set of tuples, and then feed them to the encoding layer (E-GRU: encoder-RNN).
The result of E-GRU represents the feature of each candidate network $f_{i}$. We repeat this process for all $n$ candidates and produce $f_{i}\in \{f_{1},f_{2},...,f_{n}\}$.
Then, the ranking layer, which consists of the ranking-RNN (R-GRU) and a fully-connected (FC) layer, receives the feature of a candidate network at a time and estimates the ranking score of all the candidates.
The memorization capability of the ranking layer improves the estimation accuracy since it remembers the features of the past candidate networks to estimate the relative performance of the current network.
The loss function of the SCC is defined as:
\begin{equation}
Loss_{mem}=\frac{1}{n}\sum_{i=1}^{n}((y_{i}-y_{i}^p)^2),
\label{eq:loss}
\end{equation}
Where $n$ denotes the number of input architectures, $y_{i}^p$ denotes the estimated result of candidate architecture ${i}$, and $y_{i}$ denotes the memory-efficiency metric.

We devise the memory-efficiency metric $y_i$ to compare each of the candidates in the current search round to the neural network chosen in the previous search round. It is thus formulated to:
\begin{equation}
\begin{split}
y_{i}=&\lambda\frac{a_{i}-a_{pre}}{a_{pre}}\\
&+(1-\lambda)(\frac{r_{i}-r_{pre}}{r_{pre}}+\frac{p_{i}-p_{pre}}{p_{pre}}),
\end{split}
\label{eq:metric}
\end{equation}
where $a$ is the accuracy of a neural network, $r$ is the maximum memory requirement for buffering intermediate representations, and $p$ is that for storing parameters. The subscript $pre$ denotes the neural network selected in the previous search round (i.e., the base network of the current search round) and the subscript $i$ denotes the $i-th$ candidate in the current search round. $\lambda$ is a user-specified hyper-parameter to set the target trade-off between inference network performance and memory requirement. $\lambda=0$ makes MemNAS solely aiming to minimize the memory requirement of an inference network, whereas $\lambda=1$ solely to maximize the accuracy performance.

\begin{table}[t]
\centering
\caption{\textbf{CIFAR-10 Result Comparisons.} MemNAS ($\lambda=0.5$) and ($\lambda=0.8$) are different search results with different search target trade-off between performance and memory requirement. \textit{Total Memory}: memory requirement containing the parameters memory and the essential intermediate representation memory calculated by our lifetime-based method. \textit{Memory Savings}: the savings in total memory requirement calculated by MemNAS ($\lambda=0.5$). \textit{Top-1 Acc.}: the top-1 classification accuracy on the CIFAR-10.
}
 \vspace{-0.1cm}
\begin{tabular}{p{85pt}|p{36pt}<{\centering}|p{36pt}<{\centering}|p{39pt}<{\centering}}
\hline
\specialrule{0em}{1pt}{1pt}
\multirow{2}{*}{\textbf{Model}}& \textbf{Total}&\textbf{Memory}& \textbf{Top-1} \\
&\textbf{Memory} & \textbf{Savings}&\textbf{Acc. (\%)}\\
\hline
\specialrule{0em}{1pt}{1pt}
MobileNet-V2~\cite{howard2017mobilenets}
&16.3 MB&60.7 \%&94.1\\

ResNet-110~\cite{he2016deep}
&9.9 MB&   41.1 \%&   93.5\\
ResNet-56~\cite{he2016deep}
&6.7 MB&   12.4 \%&   93.0\\
ShuffleNet~\cite{zhang2018shufflenet}
&8.3 MB&30.1 \%&92.2\\
CondenseNet-86\cite{huang2018condensenet}
&8.1 MB& 21.0 \%&   94.9 \\
CondenseNet-50\cite{huang2018condensenet}
&6.8 MB& 14.7 \%&  93.7\\
\specialrule{0em}{1pt}{1pt}
\hline
\specialrule{0em}{1pt}{1pt}
DPPNet-P~\cite{dong2018ppp-net:}
&8.1 MB&  28.4 \%& 95.3\\
DPPNet-M~\cite{dong2018ppp-net:}
&7.7 MB&   24.7 \%& 94.1\\


\specialrule{0em}{1pt}{1pt}
\hline
\hline
\specialrule{0em}{1pt}{1pt}
\textbf{MemNAS ($\lambda=0.5$)}& \textbf{5.2 MB}&$-$& 94.0\\
\textbf{MemNAS ($\lambda=0.8$)}&  6.4 MB&$-$& \textbf{95.7 }\\
\hline
\end{tabular}
\label{tab:result}
\vspace{-0.2cm}
\end{table}

\begin{table*}[t]
\centering
\caption{\textbf{ImageNet Result Comparisons.} For baseline models, we divide them into two categories according to their target trade-offs between accuracy and memory consumption. For our models, MemNAS-A and -B are extended from search models MemNAS ($\lambda=0.5$) and ($\lambda=0.8$), respectively have 16 blocks. \textit{Top-1 Acc.}: the top-1 classification accuracy on the ImageNet. \textit{Inference Latency} is measured on a Pixel Phone with batch size 1.}
 \vspace{-0.1cm}
\begin{tabular}{p{150pt}|p{38pt}<{\centering}|p{55pt}<{\centering}p{54pt}<{\centering}p{48pt}<{\centering}|p{55pt}<{\centering}}
\hline
\specialrule{0em}{1pt}{1pt}
\multirow{2}{*}{\textbf{Model}}&\multirow{2}{*}{\textbf{Type}}& \textbf{Total} &\textbf{Memory}& \textbf{Inference}&\textbf{Top-1}\\
&& \textbf{Memory} &\textbf{Savings}& \textbf{Latency}&\textbf{Acc. (\%)}\\
\hline
\specialrule{0em}{1pt}{1pt}
CondenseNet (G=C=8)~\cite{huang2018condensenet}&manual
& 24.4 MB& 6.6 \%&$-$&71.0 \\
ShuffleNet V1 (1.5x)~\cite{zhang2018shufflenet}&manual
&25.1 MB&9.2 \%&$-$&71.5\\
MobileNet V2 (1.0x)~\cite{sandler2018mobilenetv2}&manual
&33.1 MB&31.1 \%& 75 ms&71.8\\

ShuffleNet V2 (1.5x)~\cite{ma2018shufflenetv2}&manual
&26.1 MB&12.6 \%&$-$&72.6 \\

DVOLER-C~\cite{michel2019dvolver}&auto
&29.5 MB&22.7 \%&$-$&70.2 \\
EMNAS~\cite{elsken2018efficient}&auto
&54.0 MB&57.8 \%&$-$&71.7 \\
FBNet-A~\cite{wu2019fbnet}
&auto
&29.0 MB&21.4 \%&$-$&73.0 \\
MnasNet (DM=0.75)~\cite{tan2019mnasnet}
&auto
&27.4 MB&16.8 \%&61 ms&73.3 \\

DARTS~\cite{liu2018darts}&auto
&31.0 MB&38.7 \%&$-$&73.3 \\

NASNet (Mobile)~\cite{zoph2018learning}&auto
&53.2 MB&57.1 \%&183 ms&73.5 \\


\textbf{MemNAS-A (ours)}&auto
&\textbf{22.8} MB&$-$&\textbf{58 ms}& \textbf{74.1 } \\
\specialrule{0em}{1pt}{1pt}
\hline
\hline
\specialrule{0em}{1pt}{1pt}
CondenseNet (G=C=4)~\cite{huang2018condensenet}&manual
&31.6 MB&11.1 \%&$-$&73.8 \\
ShuffleNet V1 (2.0x)~\cite{zhang2018shufflenet}&manual
&33.5 MB&16.1 \%&$-$&73.7\\
MobileNet V2 (1.4x)~\cite{sandler2018mobilenetv2}&manual
&48.5 MB&42.1 \%& $-$&74.7 \\
ShuffleNet V2 (2.0x)~\cite{ma2018shufflenetv2}&manual
&51.6 MB&45.5 \%&$-$&74.9\\
DPPNet-P (PPPNet)~\cite{dong2018ppp-net:}&auto
&34.7 MB&19.0 \%&$-$&74.0 \\
ProxylessNAS-M~\cite{cai2018proxylessnas}&auto
&36.2 MB&22.4 \%& 78 ms&74.6 \\
DVOLER-A~\cite{michel2019dvolver}&auto
&39.1 MB&28.1 \%&$-$&74.8 \\
FBNet-C~\cite{wu2019fbnet}
&auto
&35.2 MB&20.2 \%&$-$&74.9 \\
MnasNet (DM=1)~\cite{tan2019mnasnet}
&auto
&36.7 MB&23.4 \%&78 ms&75.2 \\



\textbf{MemNAS-B (ours)}&auto
& \textbf{28.1 MB}&$-$& \textbf{69 ms}& \textbf{75.4 }\\
\hline
\end{tabular}
\label{tab:result-img}
\vspace{-0.2cm}
\end{table*}
\subsection{Candidate Selection}
\label{sec:am}

After the SCC produces the top-$k$ list of the candidates, MemNAS trained those candidates using the target dataset and loss function. In this work, we used the CIFAR-10 and ImageNet datasets for classification and therefore used the cross-entropy loss function, $L_{CE}$, in training candidates. We then calculate the memory-efficiency metric of each candidate with the actual accuracy performance and calculated memory requirement and re-rank them. Then, we choose the candidate with the highest-ranking score. 

We conclude the current round of MemNAS by training the SCC. Here, we use the data of the top-$k$ candidates that we just trained and their memory-efficiency metrics that we just calculated.
We used the loss function, $Loss_{mem}$, defined above in Equation~\ref{eq:loss}. After updating the SCC, we start the new round of the search if the completion criteria have not been met.

\subsubsection{Memory Requirement Estimation}
In each search round, MemNAS calculates and uses the memory-efficiency metric ($y_i$) in multiple steps namely, to estimate the top-$k$ candidates with the SCC, to train the SCC, and to determine the best candidate at the end of a search round.
As shown in Equation~\ref{eq:metric}, the metric is a function of the memory requirements for parameters and intermediate representations. It is straightforward to estimate the memory requirement of parameters. For example, we can simply calculate the product of the number of weights and the data size per weight (e.g., 2 Bytes for a short integer number). However, it is not simple to estimate the memory requirement for intermediate representations since those data are stored and discarded in a more complex manner in the course of an inference operation.
The dynamics also depend on the hardware architecture such as the size of a register file, that of on-chip data memory, and a caching mechanism, etc.

Our goal is therefore to estimate the memory requirement for buffering intermediate representations yet without the details of the underlying computing hardware architecture. To do so, we leverage a so-called register lifetime estimation technique \cite{285240}, where the lifetime of data is defined as the period from generation to deletion.
\begin{figure*}[t]
\centering
\includegraphics[width=1\linewidth]{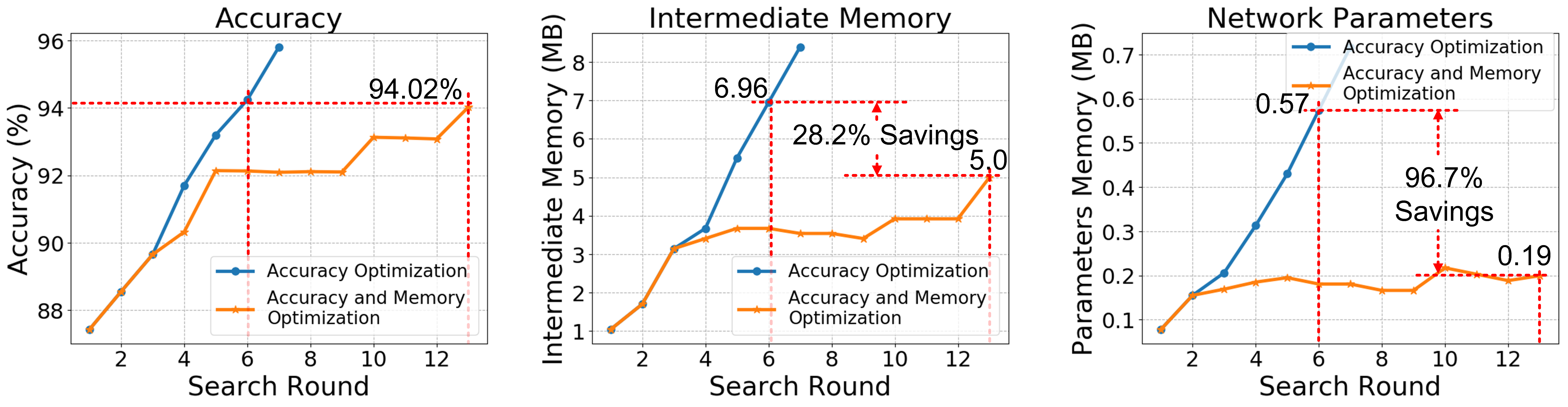}
\vspace{-0.5cm}
\caption{
\textbf{Performance and Memory Requirement over MemNAS Search Rounds on CIFAR-10.} One MemNAS is configured to optimize only accuracy performance (blue lines) and the other MemNAS to optimize both performance and memory requirement (orange line). The latter achieves the same target accuracy (94.02\%) while savings the memory requirement for parameters by 96.7\% and intermediate representation data by 28.2\%.}
\label{fig:rs}
\vspace{-0.3cm}
\end{figure*}

To perform an inference operation with a feed-forward neural network, a computing platform calculates each layer's output from the input layer to the output layer of the network. 
The outputs of a layer must be stored in the memory until it is used by all the subsequent layers requiring that. After its lifetime, the data will be discarded, which makes the memory hardware used to store those data to be available again for other data.

For the neural network shown in Figure~\ref{fig:memnet}(b), we draw the example lifetime plot (Figure~\ref{fig:lt}). In the vertical axis, we list all the edges of a neural network, i.e., intermediate representations. 
The horizontal axis represents time (T), where we assume one layer computation takes one unit time (u.t.). At T=1 u.t., $r_{1}$ is generated and stored and fed to three layers (Dep $3\times3$, Fac $1\times7$, and Dep $5\times5$). Assuming the data size of $r_1$ is 1, the memory requirement at T=1 u.t. is 1. At T=2 u.t., the three layers complete the computations and generate $I_2$, $I_3$, and $I_5$. These data indeed need to be stored. However, $r_1$ is no longer needed by any layers and thus can be discarded. Thus, at T=2 u.t., the size of the required memory is 3. We can continue this process to the last layer of the network and complete the lifetime plot. The last step is to simply find the largest memory requirement over time, which is 4 in this example case.
\section{Experiments}
\subsection{Experiment Setup}
We first had MemNAS to find the optimal neural network architecture for CIFAR-10, which contains 50,000 training images and 10,000 test images. We use the standard data pre-processing and augmentation techniques: namely
the channel normalization, the central padding of training images to 40$\times$40 and then random cropping back to 32$\times$32, random horizontal flipping, and cut-out.
The neural network architecture considered here has a total of five blocks. The number of filters in each operation layer is 64. The size of the hidden stages in the GRU model of the SCC is 100. The size of the embedding layer is also 100. The SCC estimates top-100 candidates, i.e., $k=100$.
The top-100 candidate networks are trained using Stochastic Gradient Descent with Warm Restarts \cite{loshchilov2016sgdr}. Here, the batch size is 128, the learning rate is 0.01, and the momentum is 0.9. They are trained for 60 epochs. In the end, our grow-and-trim based search process cost around 14 days with 4 GTX 1080 GPUs.

We then consider the ImageNet, which comprises 1000 visual classes and contains a total of 1.2 million training images and 50,000 validation images. Here, we use the same block architecture that MemNAS found in the CIFAR-10, but extends the number of the blocks and the filters of the inference neural network to 16 and 256 for MemNAS-A and MemNAS-B in Table~\ref{tab:result-img}.
Then, we adopt the same standard pre-processing and augmentation techniques and perform a re-scaling to 256$\times$256 followed by a 224$\times$224 center crop at test time before feeding the input image into the networks.

\begin{figure}[htb]
\centering
	\includegraphics[width=1\linewidth]{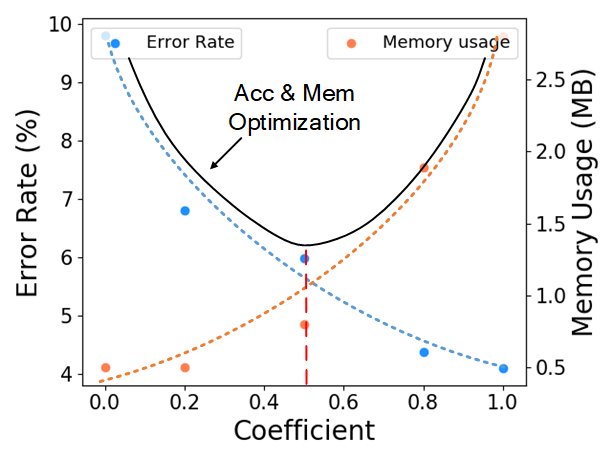}
	\vspace{-0.6cm}
	\caption{\textbf{Results on $\lambda$ Modulations.} Sweeping $\lambda$ from 0 to 1, MemNAS can produce a range of neural network architectures with well-scaling accuracy and memory requirement. The mid-value, 0.5 enables MemNAS to produce a well-balanced neural network architecture. The experiments use the CIFAR-10 data set. }
\vspace{-0.2cm}
\label{fig:trad}
\end{figure}
\subsection{Results on CIFAR-10}
Our experiment results are summarized in Table~\ref{tab:result}. We had MemNAS to construct two searched models, MemNAS ($\lambda=0.5$) and MemNAS ($\lambda=0.8$), for different target trade-offs between accuracy and memory consumption. We compared our searched models with state-of-the-art efficient models both designed manually and automatically. The primary metrics we care about are memory consumption and accuracy.

Table~\ref{tab:result} divides the approaches into two categories according to their type. Compared with manually models, our MemNAS ($\lambda=0.5$) achieves a competitive 94.0\% accuracy, better than ResNet-110 (relative +0.5\%), ShuffleNet (relative +1.8\%), and CondenseNet-50 (relative +0.3\%). Regarding memory consumption, our MemNAS ($\lambda=0.5$) saves average 27.8\% memory requirement. Specifically, our MemNAS ($\lambda=0.5$) only requires 0.8 MB parameter memory, which is satisfied for the on-chip memory storage of mobile devices, such as Note 8’s Samsung Exynos 8895 CPU, Pixel 1’s Qualcomm Snapdragon 821 CPU, iPhone 6’s Apple A8 CPU, etc.
Compared with automatic approaches, our MemNAS ($\lambda=0.8$) achieves the highest 95.7\% top-1 accuracy, while costing 8.6\% less memory requirement. The searched architectures and more comparison results are provided in the supplement material.

\subsection{Results on ImageNet}
Our experiment results are summarized in Table~\ref{tab:result-img}. We compare our searched models with state-of-the-art efficient models both designed manually and automatically. Besides, we measure the inference latency of our MemNAS-A and -B when they are performed on Pixel 1. The primary metrics we care about are memory consumption, inference latency, and accuracy.

Table~\ref{tab:result-img} divides the models into two categories according to their target trade-offs between accuracy and memory consumption. In the first group, our MemNAS-A achieves the highest 74.1\% accuracy. Regarding memory consumption, our MemNAS-A requires 35.8\% less memory consumption than other automatic approaches, while still having an average +1.6\% improvement on accuracy. In the second group, our MemNAS-B also achieves the highest 75.4\% top-1 accuracy. Regarding memory consumption, our MemNAS-B uses 22.62\% less memory than other automatic approaches, while still having an average +0.7\% improvement on accuracy. Although we did not optimize for inference latency directly, our MemNAS-A and -B are 3 ms and 9 ms faster than the previous best approaches. More comparison results are provided in the supplement material.


\subsection{Results with $\lambda$ Modulation}
We performed a set of experiments on the impact of modulating $\lambda$ in the memory-efficiency metric (Equation~\ref{eq:metric}). $\lambda$ sets the proportion between accuracy and memory requirement in MemNAS. We first compare the MemNAS with $\lambda=1$ that tries to optimize only accuracy performance and the MemNAS with $\lambda=0.5$ that tries to optimize both accuracy and memory requirement.
Figure~\ref{fig:rs} shows the results of this experiment. The MemNAS optimizing both accuracy and memory requirement achieves the same target accuracy (relative 94.02\%) while achieving 96.7\% savings in the memory requirement for parameter data and relative 28.2\% in the memory requirement for intermediate representation data.

Besides, we extend the experiment and sweep $\lambda$ from 0 to 1 (0,0.2,0.5,0.8,1) to conduct our MemNAS on CIFAR-10. Specifically, the other experiment settings follow the same setting with the previous MemNAS ($\lambda=0.5$). Figure~\ref{fig:trad} shows the results of the experiment. As the increasing of $\lambda$, asking MemNAS to less focus on memory requirement optimization, the MemNAS achieves higher accuracy performance, reducing the error rate from 9.5\% to 4\%, indeed at the cost of larger memory requirement from 0.5 to over 2.5 MB. $\lambda=0.5$ balances the accuracy performance and memory requirement.
\subsection{Experiments on the Controller}
We compare the proposed SCC to the conventional controller used in the existing NAS work. The conventional controllers \cite{michel2019dvolver} again try to estimate the absolute score of each neural network candidate individually, and then the NAS process ranks them based on the scores. We consider two types of conventional controllers, one using one-layer RNN (GRU) and the other using two-layer RNN (GRU), the latter supposed to perform better.
The performance of the controllers is evaluated with two well-received metrics, namely normalized discounted cumulative gain (NDCG) and average precision (AP). NDCG represents the importance of the selected $k$ candidates and AP represents how many of the selected $k$ candidates by a controller do remain in the real top-$k$ candidates. We consider two $k$ values of 50 and 100 in the experiment. Table~\ref{tab:rnn} shows the results. The SCC controller outperforms the conventional controllers across the evaluation metrics and the $k$ values. As expected, the two-layer RNN improves over the one-layer RNN but cannot outperform the proposed SCC.

\begin{table}[t]
\centering
\caption{\textbf{Controller Comparisons.} The proposed SCC aims to estimate the relative ranking score, outperforming the conventional controllers that estimate the absolute score of each neural network candidate and later rank them using the scores.}
\vspace{-0.1cm}
\begin{tabular}{l|c|c|c|c}
\hline
\multirow{2}{*}{\textbf{Model}}&\textbf{AP}&\textbf{AP}&\textbf{NDCG}&\textbf{NDCG}\\
&\textbf{@50}&\textbf{@100}&\textbf{@50}&\textbf{@100}\\
\hline
Single-RNN&0.066&0.237&0.043&0.078\\
Double-RNN&0.128&0.268&0.062&0.080\\
\textbf{Our Method}&\textbf{0.196}&\textbf{0.283}&\textbf{0.135}&\textbf{0.201}\\
\hline
\end{tabular}
\label{tab:rnn}
\vspace{-0.3cm}
\end{table}
\section{Conclusion}
In this work, we propose MemNAS, a novel NAS technique that can optimize both accuracy performance and memory requirement. The memory requirement includes the memory for both network parameters and intermediate representations. We propose a new candidate generation technique that not only grows but also trims the base network in each search round, and thereby increasing the size and the diversity of the search space.
To effectively find the best architectures in the search space, we propose the structure correlation controller, which estimates the ranking of candidates by the relative information among the architectures.
On the CIFAR-10, our MemNAS achieves 94\% top-1 accuracy, similar with  MobileNetV2 (94.1\%)~\cite{sandler2018mobilenetv2}, while only requires less than 1 MB parameter memory.
On the ImageNet, our MemNAS achieves 75.4\% top-1 accuracy, 0.7\% higher than MobileNetV2~\cite{sandler2018mobilenetv2} with 42.1\% less memory requirement. 

{\small

\bibliographystyle{ieeetr}

}

\end{document}